\documentclass{article}

\usepackage{amsmath}
\usepackage{amssymb}
\usepackage{booktabs}
\usepackage{float}
\usepackage{graphicx}
\usepackage{tabularx}
\usepackage{longtable}
\usepackage{array}
\usepackage{placeins}
\usepackage{algpseudocode}
\usepackage[preprint]{corl_2026} 

\newcolumntype{Y}{>{\raggedright\arraybackslash}X}
\floatstyle{ruled}
\newfloat{algorithm}{tbp}{loa}
\floatname{algorithm}{Algorithm}
\algrenewcommand\algorithmicrequire{\textbf{Input:}}
\algrenewcommand\algorithmicensure{\textbf{Output:}}
\makeatletter
\newcommand{\compactappendixspacing}{%
  \raggedbottom
  \setlength{\parskip}{2pt}
  \setlength{\textfloatsep}{5pt}
  \setlength{\floatsep}{5pt}
  \setlength{\intextsep}{5pt}
  \setlength{\dbltextfloatsep}{5pt}
  \setlength{\dblfloatsep}{5pt}
  \setlength{\abovecaptionskip}{3pt}
  \setlength{\belowcaptionskip}{3pt}
  \setlength{\@nipsabovecaptionskip}{3pt}
  \setlength{\@nipsbelowcaptionskip}{3pt}
  \setlength{\abovedisplayskip}{4pt plus 1pt minus 2pt}
  \setlength{\belowdisplayskip}{4pt plus 1pt minus 2pt}
  \setlength{\abovedisplayshortskip}{3pt plus 1pt minus 1pt}
  \setlength{\belowdisplayshortskip}{3pt plus 1pt minus 1pt}
  \renewcommand{\section}{%
    \@startsection{section}{1}{\z@}%
                  {-1.2ex \@plus -0.2ex \@minus -0.1ex}%
                  {0.7ex \@plus 0.1ex}%
                  {\large\bf\raggedright}%
  }%
  \renewcommand{\subsection}{%
    \@startsection{subsection}{2}{\z@}%
                  {-1.0ex \@plus -0.2ex \@minus -0.1ex}%
                  {0.5ex \@plus 0.1ex}%
                  {\normalsize\bf\raggedright}%
  }%
  \renewcommand{\paragraph}{%
    \@startsection{paragraph}{4}{\z@}%
                  {0.7ex \@plus 0.1ex \@minus 0.1ex}%
                  {-1em}%
                  {\normalsize\bf}%
  }%
}
\makeatother

\title{HEFT: Heavy-Payload Full-size Humanoid Teleoperation with Privileged Motion Guidance and Windowed Payload Curriculum}

\author{
  {\bfseries
  Chenxin Liu\textsuperscript{1,2,*}\quad
  Qingzhou Lu\textsuperscript{1,2,*}\quad
  Guangxiao Yang\textsuperscript{2}\quad
  Xuanyang Shi\textsuperscript{2}}\\
  {\bfseries
  Chenghan Yang\textsuperscript{1}\quad
  Yanjiang Guo\textsuperscript{1,2}\quad
  Jianyu Chen\textsuperscript{1,2,3\dag}}\\[0.4em]
  {\normalfont\textsuperscript{1}Tsinghua University\quad
  \textsuperscript{2}RobotEra\quad
  \textsuperscript{3}Shanghai Qizhi Institute}\\[0.3em]
  {\normalfont\textsuperscript{*}Equal contribution.\quad
  \textsuperscript{\dag}Corresponding author.}
}

\begin{document}
\vspace*{-0.5in}
\maketitle

\vspace{-2.1em}
\noindent\makebox[\linewidth][c]{%
  \normalfont\small\textbf{Project Page:} \href{https://heft.axell.top/}{\texttt{HEFT-homepage}}%
}\par
\vspace{0.55em}
\begin{figure}[h]
  \centering
  \makebox[\linewidth][c]{\hspace*{-0.83cm}\includegraphics[width=0.88\linewidth]{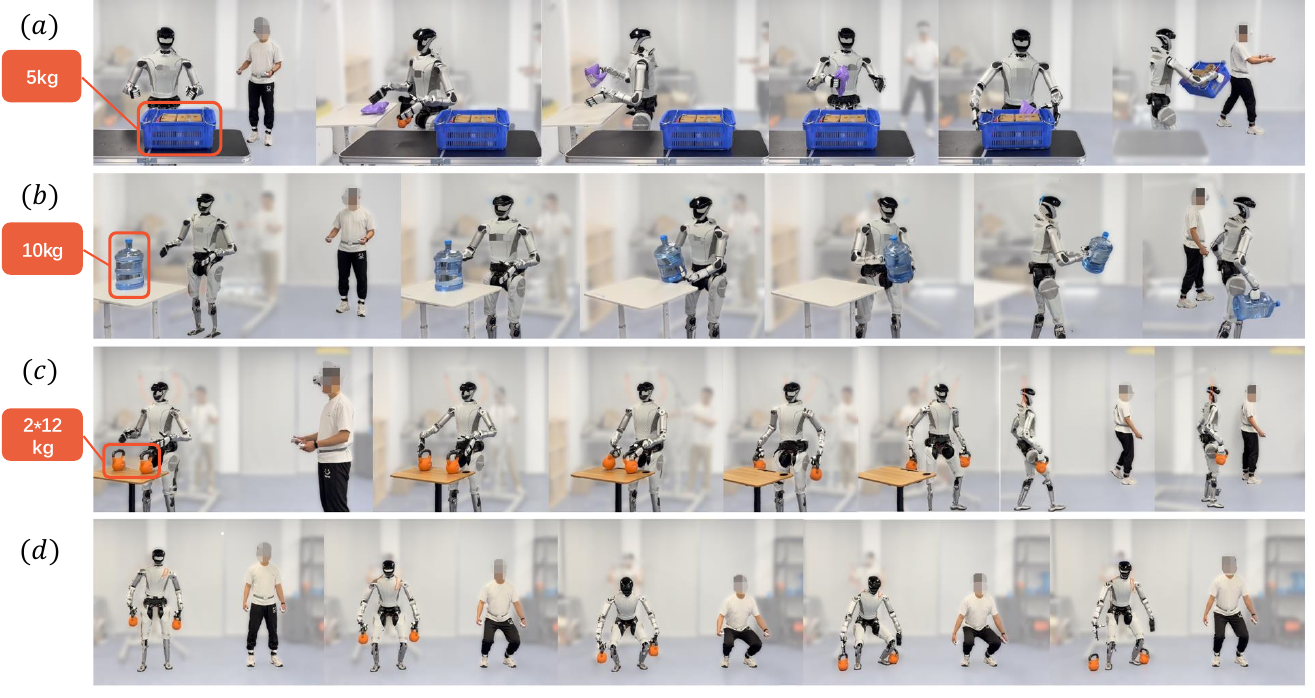}}
  \caption{Heavy-payload teleoperation on the full-size L7 humanoid using the same deployable policy: (a) picking up and carrying a 5\,kg loaded basket after multi-stage tabletop handling, (b) picking up and asymmetrically carrying a 10\,kg water bottle, (c) walking with two 12\,kg kettlebells, and (d) squatting with the same 24\,kg total payload.}
  \label{fig:teaser}
\end{figure}

\begin{abstract}
General motion tracking and teleoperation offer a promising path to scalable humanoid skill acquisition, yet most existing frameworks are validated on compact platforms or without real payload interaction, leaving full-size humanoids with real payloads largely unexplored. Scaling to full-size humanoids introduces two compounding challenges: their larger inertia and tighter balance margins make tracking highly sensitive to noise, drift, and retargeting errors from commodity VR trackers, while their payload potential remains largely underutilized. We present HEFT, a heavy-payload full-size humanoid teleoperation framework that addresses both challenges. HEFT learns from deployable noisy VR references with physically plausible reconstructed references through Privileged Motion Guidance (PMG), and uses a Windowed Payload Curriculum (WPC) with expert-guided payload caps to acquire robust heavy-payload tracking. We deploy HEFT on L7, a 175\,cm, 65\,kg humanoid. The robot tracks motions including turns, forward/backward locomotion, and squats under payloads up to 24\,kg.
\end{abstract}


\section{Introduction}

Whole-body teleoperation is becoming a practical interface for scaling humanoid skill acquisition: a human operator supplies task intent, timing, object interactions, and motion style, while the robot controller converts human references into dynamically feasible actions. Teleoperated demonstrations have already enabled scalable learning for manipulation and mobile manipulation~\cite{mandlekar2018roboturk,zhao2023learning,fu2024mobile,wu2024gello}, and recent humanoid systems have extended this idea to immersive teleoperation, shadowing, whole-body imitation, and demonstration collection~\cite{cheng2024open,he2024learning,he2024omnih2o,fu2024humanplus,ze2025twist,ze2025twist2}. Most existing validations, however, focus on compact platforms or motions without real payload interaction. Full-size humanoids are designed to operate at human scale, where useful tasks often involve carrying and manipulating human-scale objects. This motivates studying payload-bearing teleoperation at human scale.

This setting raises two challenges that are easy to hide in smaller-scale evaluations or settings without real payload interaction. First, deployable VR tracking provides a noisy online reference rather than a clean motion-capture trajectory. VR observations capture only sparse head, hand, and foot measurements, making full-body motion recovery underconstrained~\cite{jiang2022avatarposer,ponton2023sparseposer,winkler2022questsim,li2023ego}. In our setup, the raw reference stream can further exhibit drift, latency, body-frame misalignment, and biased limb positions. Offline human-motion reconstruction can improve motion plausibility and global consistency from imperfect observations~\cite{zhang2024rohm,rempe2021humor,shin2024wham,wang2024tram,shen2024world}, but treating reconstructed motion as the online robot command would add delay and weaken the operator's feedback loop. The deployed controller therefore has to act from the raw VR stream. However, full-size humanoid teleoperation is more sensitive to VR artifacts. On a full-size robot, small upper-body reference errors may induce large centroidal momentum and contact disturbances.

Second, payloads do not impose a single uniform difficulty across a teleoperated motion. The same load may be manageable during quasi-static support but destabilizing during turns, squats, or fast arm swings, so object forces must be handled jointly with balance, contact, and posture~\cite{dao2024sim,zhang2025falcon,purushottam2025heavy,wang2026halo}. Existing methods, however, often target specific tasks, platforms, or prescribed force settings, which are difficult to scale to broad whole-body reference libraries and can degrade performance on complex motions. This calls for a payload curriculum that adapts to motion-dependent feasibility while remaining compatible with broad whole-body tracking.

We present HEFT, a heavy-payload full-size humanoid teleoperation framework that addresses these two challenges with training-time guidance. For noisy VR references, HEFT trains the deployable policy on raw VR inputs while using physically plausible reconstructed references through Privileged Motion Guidance (PMG). For payloads, HEFT uses a Windowed Payload Curriculum (WPC): it labels short motion windows with expert-guided feasible payload caps and samples the two-hand load within each cap. We deploy HEFT on L7, a 175\,cm, 65\,kg humanoid with 29 actuated joints. Our method tracks dynamic motions and loaded teleoperation motions, including turns, forward/backward locomotion, and squats, with real two-hand payloads up to 24\,kg.

In summary, our contributions are:
\begin{itemize}
  \item a Privileged Motion Guidance formulation that trains from raw VR references while using reconstructed motion only during training;
  \item a Windowed Payload Curriculum that expands heavy-payload tracking without a task-specific carrying controller;
  \item a full-size L7 validation of 24\,kg payload teleoperation and dynamic tracking with a single controller.
\end{itemize}

\section{Related Work}
\label{sec:related}

\paragraph{Humanoid teleoperation and motion tracking.}
Physics-based humanoid motion tracking has progressed from example-guided imitation to reusable motion priors and general-purpose whole-body trackers~\cite{peng2018deepmimic,peng2021amp,peng2022ase,luo2023perpetual,luo2024universal,tessler2024maskedmimic,liao2025beyondmimic,yin2026unitracker}. Recent humanoid teleoperation systems connect these controllers with human sensing, head-mounted devices, cameras, and immersive feedback for real-time operation and data collection~\cite{cheng2024open,he2024learning,he2024omnih2o,fu2024humanplus,ze2025twist}. TWIST2 lowers the cost of collecting humanoid demonstrations with a portable VR-based system~\cite{ze2025twist2}, while SONIC scales whole-body tracking toward large foundation controllers with broad motion data and multiple input interfaces~\cite{luo2026sonic}. Despite this progress, full-size humanoid teleoperation with substantial physical payloads remains comparatively underexplored.

\paragraph{Motion reconstruction from sparse and noisy references.}
Deployable VR teleoperation provides only partial and noisy measurements, motivating methods that reconstruct complete human motion from sparse trackers, head-mounted or egocentric observations, and monocular video~\cite{jiang2022avatarposer,ponton2023sparseposer,winkler2022questsim,li2023ego,shin2024wham,wang2024tram,shen2024world}. These works differ in sensing assumptions, but all aim to recover unobserved body motion and more globally consistent references from imperfect inputs. Our offline cleanup mainly builds on RoHM~\cite{zhang2024rohm}, which trains diffusion-based denoisers on synthetically corrupted AMASS motions to recover plausible global trajectories and local body poses~\cite{mahmood2019amass}. Since such iterative reconstruction is not suitable for low-latency online control, we use it only as offline training guidance.

\paragraph{Payload-aware whole-body control.}
Carrying heavy objects turns humanoid tracking into a force-coupled whole-body control problem: payload-induced wrenches affect balance, contact timing, posture, and arm motion simultaneously. Prior loco-manipulation systems have studied learned whole-body policies for coordinating locomotion and manipulation under object interaction~\cite{fu2023deep,dao2024sim}, and recent force-aware methods further introduce explicit force adaptation, heavy-object control, haptic heavy lifting, or heavy-loaded humanoid skills~\cite{zhang2025falcon,rigo2024hierarchical,purushottam2025heavy,wang2026halo}. These works demonstrate the importance of modeling external loads, but they are often organized around specific tasks, command spaces, or prescribed force curricula. In teleoperated motion tracking, payload feasibility instead varies with the reference motion itself. The same load can be easy during static support but unstable during turns, squats, or fast upper-body motion. HEFT therefore studies payload robustness in the reference-conditioned setting, where heavy-load training must remain compatible with diverse teleoperation motions.

\section{Method}
\label{sec:method}

Fig.~\ref{fig:method_overview} summarizes HEFT. The policy tracks raw VR-derived references using proprioception and action history. During training, HEFT adds two offline signals that are unavailable at test time: a cleaner reconstructed motion paired with the raw VR stream, and motion-window payload caps estimated by expert rollouts. Sec.~\ref{sec:problem_setup} defines this reference-conditioned payload-tracking problem, Sec.~\ref{sec:asym_learning} introduces Privileged Motion Guidance, Sec.~\ref{sec:payload} introduces the Windowed Payload Curriculum, and Sec.~\ref{sec:controller_training} describes how these signals are distilled into the final controller.

\subsection{Reference-Conditioned Payload Teleoperation}
\label{sec:problem_setup}

HEFT learns a whole-body policy for tracking teleoperated reference motions under possible two-hand payloads. At time $t$, the policy observes
\begin{equation}
    o_t^{\mathrm{dep}} = \left(S_{\mathrm{prop},t}, S_{\mathrm{raw},t}\right),
\end{equation}
where $S_{\mathrm{prop},t}$ contains proprioception and action history, and $S_{\mathrm{raw},t}$ contains near-past and future targets from the raw VR reference stream. The policy outputs a joint-position command $a_t$, which is tracked by a low-level PD controller.

Each training episode samples a reference motion $m_i$ and a windowed payload schedule $F_{\mathrm{window}}$. A reference motion provides two aligned streams, $S_{\mathrm{raw}}^i$ and $S_{\mathrm{clean}}^i$. $S_{\mathrm{raw}}^i$ is the deployable command stream. $S_{\mathrm{clean}}^i$ is a training-only target used for reward, critic input, and privileged encoding. For ordinary mocap clips, the two streams are identical after retargeting. For VR clips, $S_{\mathrm{raw}}^i$ comes from the raw VR-derived motion, while $S_{\mathrm{clean}}^i$ comes from offline reconstruction and retargeting. The payload schedule applies external forces to the two wrist links during simulation. The sampled payload affects the transition dynamics, but it is not directly observed by the deployed policy. Training therefore optimizes a policy that acts from raw references while being evaluated against the cleaner target motion under payload-disturbed dynamics:
\begin{equation}
\label{eq:heft_objective}
    \max_{\pi}\;
    \mathbb{E}_{m_i,F_{\mathrm{window}}}
    \left[
    \sum_{t=0}^{T}
    \gamma^t
    R\!\left(s_t,a_t;S_{\mathrm{clean},t}^i\right)
    \right],
    \qquad
    a_t \sim \pi(\cdot \mid o_t^{\mathrm{dep}}).
\end{equation}
This objective leaves two training challenges. The policy should act from noisy raw references while being encouraged toward the cleaner reconstructed motion, and it should also learn payload robustness while respecting that different motion segments can tolerate different payload levels. PMG addresses the first challenge through privileged reconstructed references, while WPC addresses the second through motion-window payload caps.

\begin{figure}[t]
  \centering
  \includegraphics[width=\linewidth]{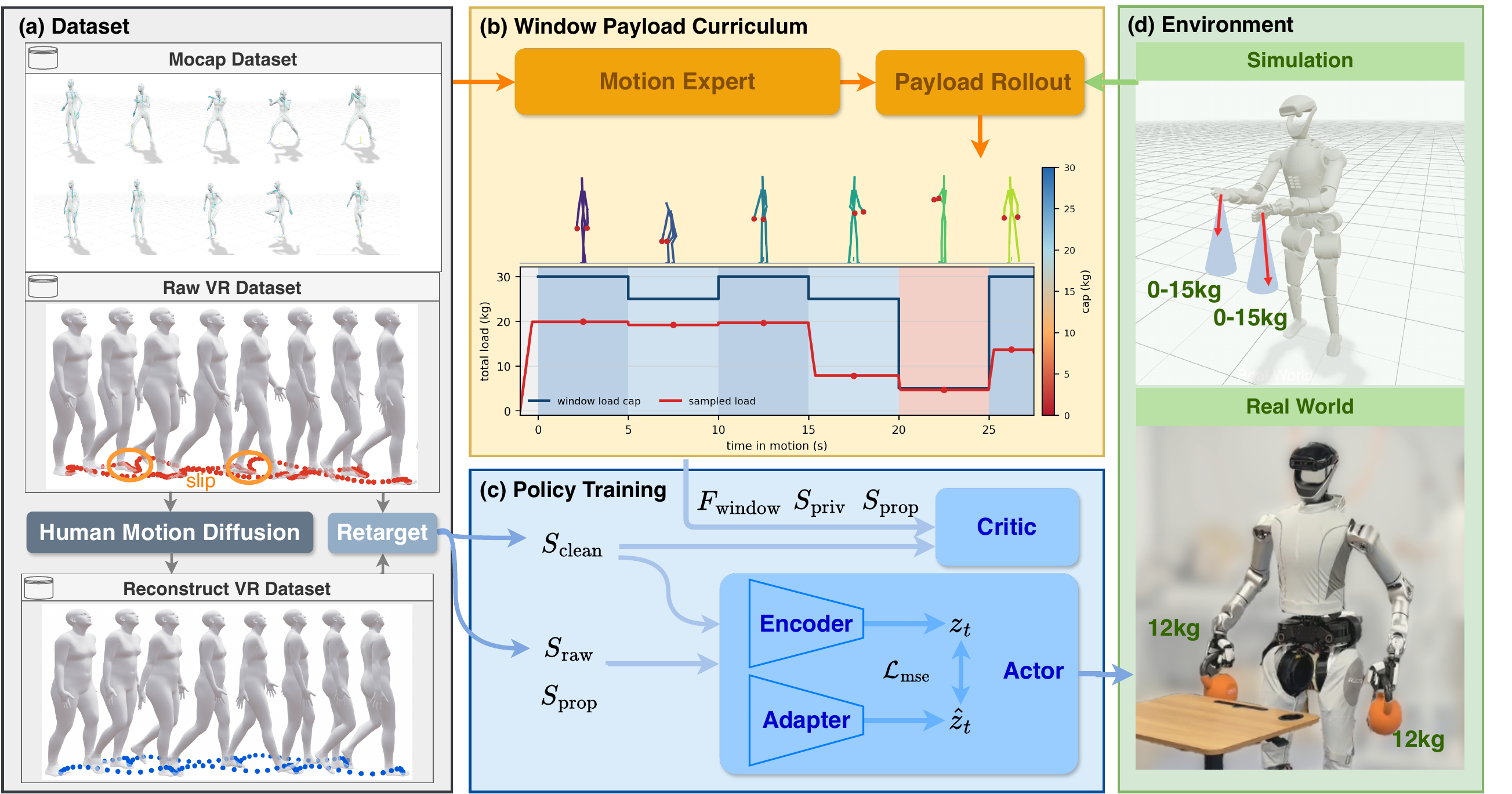}
  \caption{Method overview. (a) HEFT builds paired raw and reconstructed references from mocap and VR data; human-motion diffusion cleans structured VR artifacts before retargeting. (b) The Windowed Payload Curriculum uses a privileged motion expert and payload rollouts to estimate window-level payload caps and sample smoothed two-hand loads. (c) Policy training combines raw references and proprioception for the actor with clean references, payload state, and simulator information for privileged critic and latent adaptation. (d) The resulting controller is evaluated in simulation and deployed on L7 for real heavy-payload teleoperation.}
  \label{fig:method_overview}
\end{figure}

We build the training references from a large mocap library and a smaller paired VR set, as shown in Fig.~\ref{fig:method_overview}(a). The mocap library contains no-mirror SEED~\cite{bonesseed}, 100STYLE~\cite{mason2022real}, and LaFAN1~\cite{harvey2020robust} motions; for these clips, $S_{\mathrm{raw}}=S_{\mathrm{clean}}$. The paired VR set contains raw VR-derived references and corresponding reconstructed references. We also train an offline clean-reference tracking expert for data preparation: it filters dynamically inconsistent mocap clips and later provides the rollout labels used by WPC.

\subsection{Privileged Motion Guidance}
\label{sec:asym_learning}

PMG addresses the mismatch between the raw deployable command $S_{\mathrm{raw}}$ and the cleaner motion target $S_{\mathrm{clean}}$ constructed in Fig.~\ref{fig:method_overview}(a). VR references contain structured artifacts during locomotion, including global drift, body-frame bias, latency, and end-effector offsets. These artifacts are not well modeled by independent Gaussian noise and are not removed by robot retargeting alone. We therefore train on paired raw/reconstructed VR clips rather than relying only on mocap data or generic reference noise.
Because the VR pipeline outputs SMPL-X sequences~\cite{pavlakos2019expressive}, we can directly apply an existing human-motion reconstruction model for offline cleanup. In our implementation we use RoHM~\cite{zhang2024rohm}. The reconstructed sequence preserves the operator's style and timing while removing much of the physically implausible drift. After retargeting, each VR clip provides a pair $(S_{\mathrm{raw}},S_{\mathrm{clean}})$.

PMG inserts the paired raw/reconstructed data into the asymmetric actor-critic setup in Fig.~\ref{fig:method_overview}(c). When a paired VR clip is sampled, the actor receives reference targets from $S_{\mathrm{raw}}$. The critic and reward computation use the paired $S_{\mathrm{clean}}$ as the tracking target. For ordinary mocap clips, $S_{\mathrm{clean}}$ is simply the retargeted mocap reference and $S_{\mathrm{raw}}=S_{\mathrm{clean}}$. This separation makes the policy practice on the noisy inputs it will receive online, while anchoring rewards and value estimates to a physically plausible motion target. As a result, the controller learns which components of the raw reference reflect operator intent and which should be treated as tracker artifacts, improving global tracking without an explicit tracker-noise model.

\subsection{Windowed Payload Curriculum}
\label{sec:payload}

Fig.~\ref{fig:method_overview}(b) shows the payload pipeline. A motion does not have a single payload limit: the feasible load can differ between support phases, turns, squats, jumps, and fast arm swings. We therefore partition each motion into 5\,s windows $w_{i,k}$ and assign each window an expert-guided cap $\bar F_{i,k}$.

To generate caps, we use a reference-tracking expert trained with clean references and privileged simulator information. Because this expert is not limited by the raw-VR deployment interface, its rollouts provide an information-rich estimate of the payload that the robot dynamics can tolerate for each motion segment. We roll out the expert under downward force on the two wrist-roll bodies and search each window from 30\,kg downward in 5\,kg steps. The largest successful load becomes the cap $\bar F_{i,k}\in[0,30]$\,kg.

During policy training, $F_{\mathrm{window}}$ is sampled from a continuous kg range. Let $p\in[0,1]$ be training progress. For a sampled window $w_{i,k}$, the total two-hand payload is sampled by Eq.~(\ref{eq:window_load_sample}):
\begin{equation}
\label{eq:window_load_sample}
F_{i,k}\sim\mathcal{U}\!\left(0,\bar F_{i,k}\,
\mathrm{clip}\!\left(\frac{p}{0.8},0,1\right)\right).
\end{equation}
Loads are sampled independently across windows, split between the two wrists with randomized fractions, applied within a 12$^\circ$ cone around downward gravity. We include the load state in privileged observations to accelerate policy learning.

\subsection{Deployable Controller Training}
\label{sec:controller_training}

As shown in Fig.~\ref{fig:method_overview}(c), PMG and WPC define training signals that are unavailable at deployment. We therefore train the controller with an RMA-style teacher-student structure~\cite{kumar2021rma}. The teacher uses a privileged latent $z_t=E_p(S_{\mathrm{clean},t},S_{\mathrm{priv},t})$, where $S_{\mathrm{priv},t}$ includes simulator-only information such as tracking errors, contact signals, payload state, and randomization variables. An adapter predicts the same latent space from policy-visible history, $\hat z_t=E_a(S_{\mathrm{prop},t},S_{\mathrm{raw},t})$, and is supervised by
\begin{equation}
\label{eq:adapt_loss}
    \mathcal{L}_{\mathrm{adapt}} = \|\hat z_t-z_t\|_2^2 .
\end{equation}
The teacher and student share the same proprioceptive and raw-reference inputs, but the teacher policy conditions on $z_t$ while the student policy conditions on $\hat z_t$. Thus, privileged information shapes the latent during training, and the adapter provides its deployment-time estimate.

Training then proceeds from privileged learning to deployment-compatible adaptation. We first optimize the teacher with PPO~\cite{schulman2017proximal} under PMG and WPC, then distill its latent into the adapter using Eq.~(\ref{eq:adapt_loss}), and finally fine-tune the student while rolling out the adapter-predicted latent. At hardware deployment, only the student policy and adapter are used; reconstructed references, window caps, payload states, and simulator privileged states are removed.

\section{Experiments}
\label{sec:experiments}

\subsection{Setup}
\label{sec:exp_setup}

We evaluate HEFT in both simulation and hardware experiments. For fair comparison with existing baselines, we conduct simulation experiments on both G1 and the full-size L7 robot, and evaluate real-world deployment on L7 hardware. Our evaluation focuses on two questions. First, does Privileged Motion Guidance (PMG) improve adaptation to noisy VR data without sacrificing general tracking performance? Second, does expert-guided Windowed Payload Curriculum (WPC) improve high-payload capability while largely preserving unloaded motion tracking?

\textbf{Evaluation Datasets.} We prepare three evaluation datasets. First, we record eight VR trajectories, denoted as $\mathcal{D}_{\mathrm{VR}}$, covering locomotion at different speeds and in different directions. After retargeting, each motion corresponds to approximately 5\,m of travel on L7 and 3\,m on G1. These VR motions contain real tracking artifacts and are used to evaluate adaptation to noisy inputs. Second, based on motion statistics, we select high-dynamic motions from the SEED dataset to form $\mathcal{D}_{\mathrm{dynamic}}$. Third, we uniformly sample another 100 SEED motions to form $\mathcal{D}_{\mathrm{random}}$.

\textbf{Metrics.} We report success rate, final horizontal root tracking error, mean per-joint position error (MPJPE), body-frame root linear velocity error, and body-frame root angular velocity error. Success rate is the percentage of successful rollouts. Final horizontal root error is computed at the end of each trajectory as the L2 error, in meters, between the achieved and reference root displacement in the horizontal $xy$ plane relative to the initial state. MPJPE is reported in meters and averaged over valid rollout frames, joints, and trajectories. The root linear and angular velocity errors are computed in the body frame as framewise L2 errors averaged over valid rollout frames and trajectories, reported in m/s and rad/s respectively.

\subsection{Privileged Motion Guidance}
\label{sec:exp_pmg}

\textbf{Evaluation.} To test whether PMG improves adaptation to noisy VR inputs without degrading general tracking, we compare it on both G1 and L7 against two mocap-only ablations, one trained without motion noise (w/o noise) and one trained with motion noise (w/ noise), as well as two representative tracking baselines, SONIC and TWIST2. SONIC and TWIST2 are evaluated on G1 using the official checkpoints. On L7, we adapt TWIST2 to our stack; due to compute limits, we do not migrate SONIC to L7 at this stage. Both evaluations are conducted under the no-payload setting. On $\mathcal{D}_{\mathrm{VR}}$, where structured VR noise makes local tracking metrics unreliable, we mainly evaluate long-horizon root tracking error. On $\mathcal{D}_{\mathrm{random}}$, we evaluate success rate and tracking accuracy under noise-free reference inputs.

\begin{figure}[H]
\centering
\includegraphics[width=\linewidth]{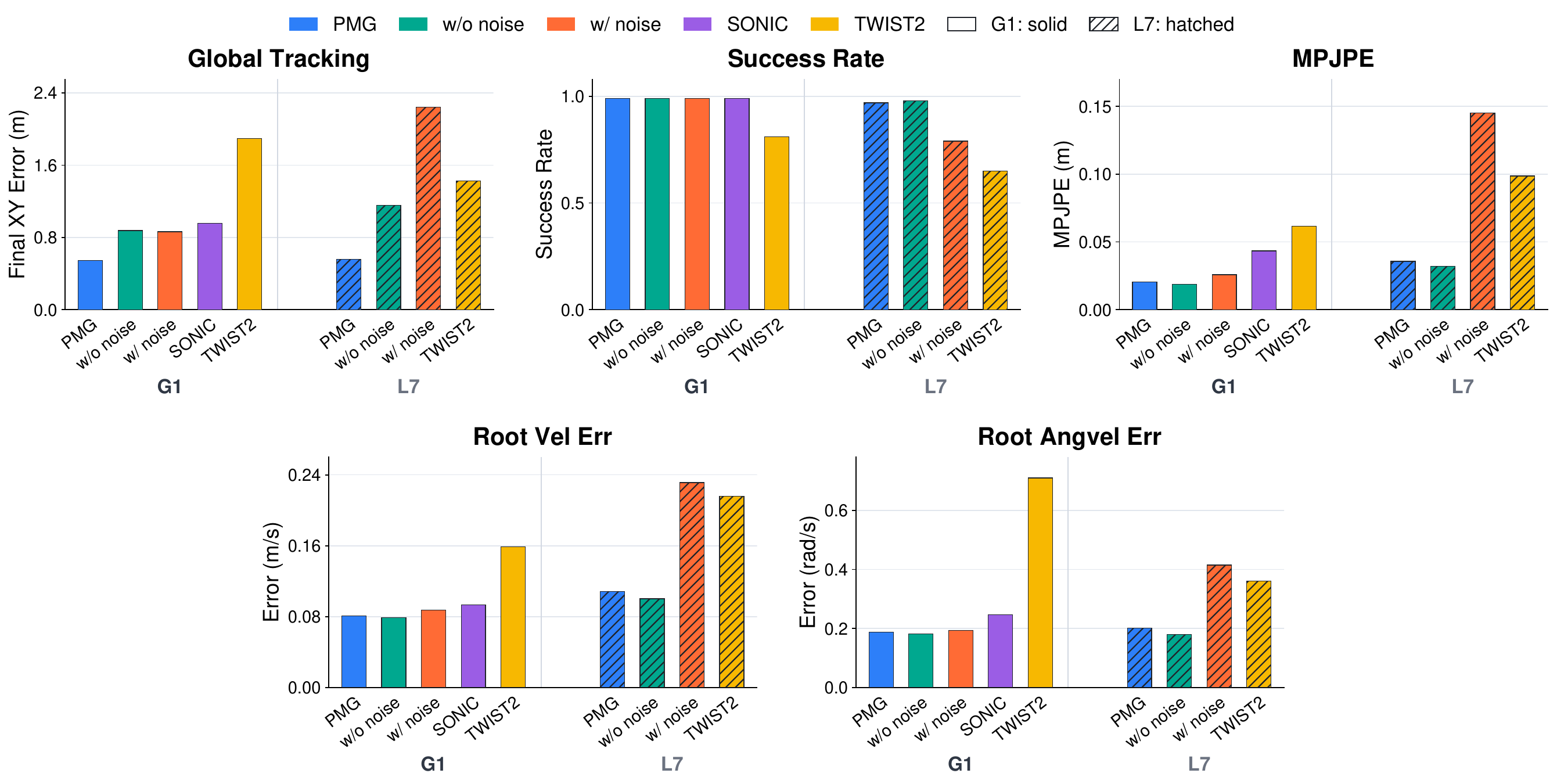}
\caption{PMG tracking comparison on G1 and L7. On noisy VR motions in $\mathcal{D}_{\mathrm{VR}}$, PMG improves long-horizon root tracking. On noise-free motions in $\mathcal{D}_{\mathrm{random}}$, PMG stays close to clean-reference training while comparing favorably with representative recent tracking baselines.}
\label{fig:pmg_tracking}
\end{figure}

\textbf{Results.} Fig.~\ref{fig:pmg_tracking} shows that PMG improves noisy-VR adaptation while largely preserving general tracking performance. On $\mathcal{D}_{\mathrm{VR}}$, PMG achieves the lowest final horizontal root error on both robots (0.544\,m on G1 and 0.560\,m on L7), reducing long-horizon drift compared with mocap-only training, generic motion-noise training, and TWIST2. On $\mathcal{D}_{\mathrm{random}}$, PMG remains close to clean-reference training, with high success on both robots and only a small MPJPE increase. It also compares favorably with recent tracking baselines on pose tracking: MPJPE is 0.021\,m for PMG versus 0.043\,m for SONIC and 0.061\,m for TWIST2 on G1, and 0.036\,m for PMG versus 0.099\,m for TWIST2 on L7. Together, these results suggest that PMG learns to interpret noisy VR references through paired raw/reconstructed data. It suppresses structured VR artifacts that harm long-horizon root tracking, while retaining the motion-tracking accuracy of clean-reference training.

\subsection{Windowed Payload Curriculum}
\label{sec:exp_window_load}

\textbf{Evaluation.} To test whether WPC improves high-payload tracking while largely preserving unloaded motion tracking, we compare with TWIST2+FC and ablate the expert-guided caps in WPC on L7. FALCON is a payload-aware baseline that introduces a force curriculum, but it is not a whole-body tracking policy. We therefore construct TWIST2+FC by transferring the FALCON-style force curriculum to the TWIST2 tracking framework, making it a comparable payload-curriculum baseline. During evaluation, load denotes the total two-hand payload and is split evenly between the hands. We use $\mathcal{D}_{\mathrm{random}}$ to evaluate tracking capability under different payload levels, and use $\mathcal{D}_{\mathrm{dynamic}}$ under the no-payload setting to test whether WPC preserves tracking performance on difficult high-dynamic motions.

\begin{figure}[H]
\centering
\begin{minipage}[t]{\linewidth}
\centering
\includegraphics[width=\linewidth]{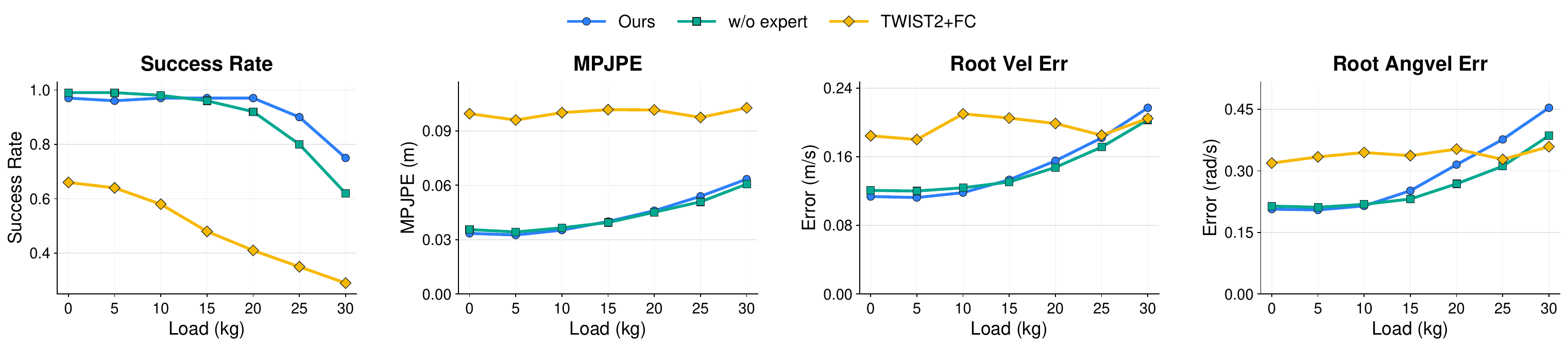}
\end{minipage}
\caption{Payload evaluation on $\mathcal{D}_{\mathrm{random}}$ under different total two-hand payloads. WPC maintains higher success at large loads than TWIST2+FC and the w/o expert ablation, while keeping pose and velocity errors close to the ablation and generally lower than TWIST2+FC.}
\label{fig:payload_load}
\end{figure}

\begin{table}[H]
\centering
\scriptsize
\caption{No-payload evaluation on $\mathcal{D}_{\mathrm{dynamic}}$.}
\label{tab:dynamic_load_ablation}
\setlength{\tabcolsep}{2.0pt}
\renewcommand{\arraystretch}{1.02}
\begin{tabular*}{\linewidth}{@{\extracolsep{\fill}}ccccc@{}}
\toprule
\textbf{Policy} & \textbf{Success $\uparrow$} & \textbf{MPJPE (m) $\downarrow$} & \textbf{Root Vel Error (m/s) $\downarrow$} & \textbf{Root Angvel Error (rad/s) $\downarrow$} \\
\midrule
w/o expert & 0.64 & 0.060 & 0.872 & \textbf{0.937} \\
Ours & \textbf{0.73} & \textbf{0.057} & \textbf{0.743} & 0.968 \\
TWIST2+FC & 0.10 & 0.133 & 1.463 & 1.294 \\
\bottomrule
\end{tabular*}
\end{table}

\textbf{Results.} Fig.~\ref{fig:payload_load} shows that WPC mainly improves success near the payload limit. Ours maintains high success up to 20\,kg and still succeeds on 90\% of trials at 25\,kg and 75\% at 30\,kg. In contrast, the w/o expert ablation drops to 80\% and 62\% at the same loads, while TWIST2+FC falls to 35\% and 29\%. The tracking metrics show a complementary trend: Ours remains close to the ablation in MPJPE and velocity errors across loads, and more accurate than TWIST2+FC in pose tracking. This suggests that expert-guided window caps primarily expand the feasible high-load regime. Table~\ref{tab:dynamic_load_ablation} further shows that WPC preserves difficult no-payload tracking, improving success from 64\% to 73\% over the w/o expert ablation while also slightly reducing MPJPE and root velocity error. TWIST2+FC remains much less stable on these high-dynamic motions and has larger tracking errors.

\subsection{Real-Robot Teleoperation}
\label{sec:exp_smooth_real}

\begin{figure}[t]
\centering
\includegraphics[width=\linewidth]{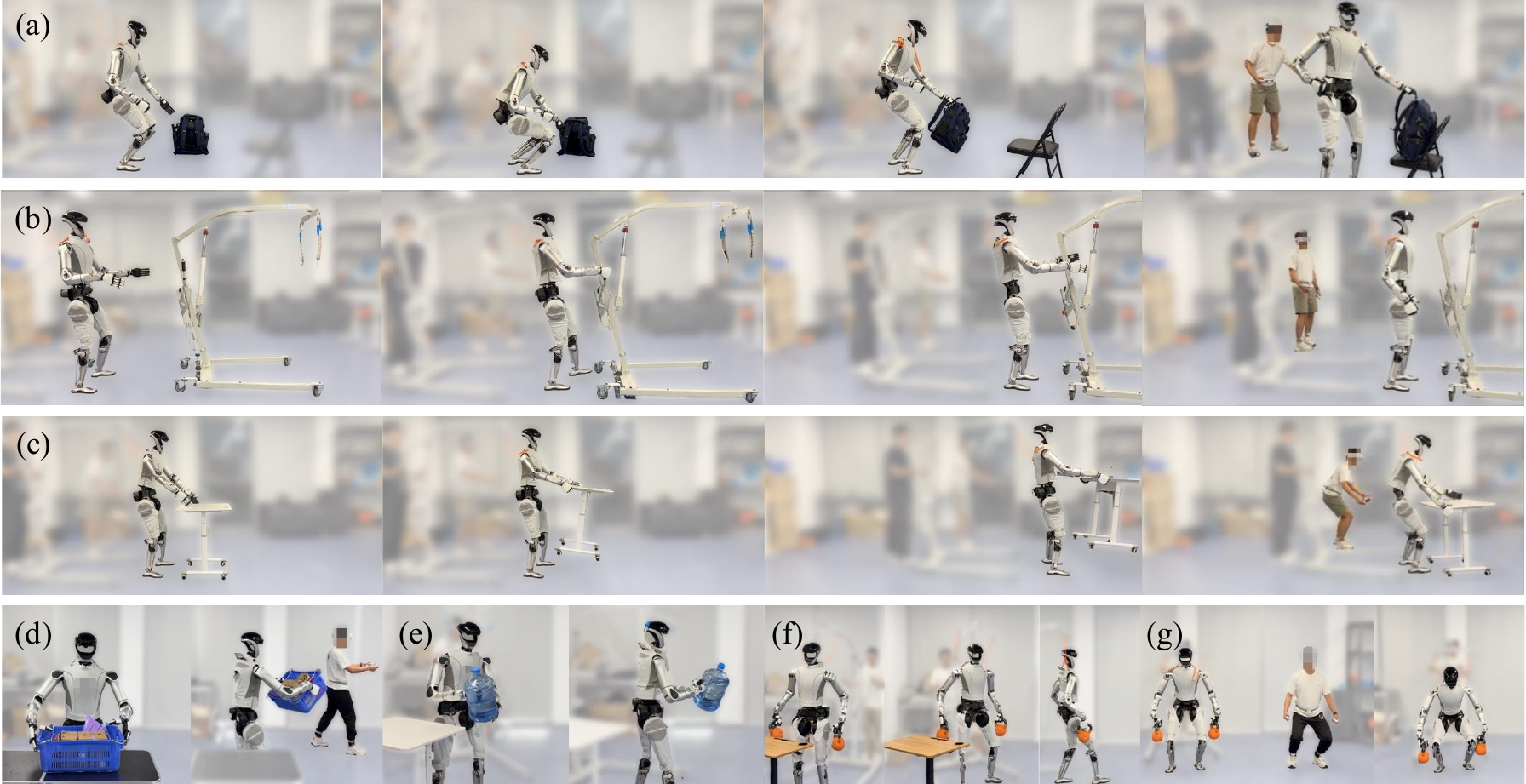}
\caption{Real-robot teleoperation tasks on L7 using the same policy. (a) Picking up a backpack from the floor and placing it on a chair, (b) pushing a wheeled rack, (c) lifting and carrying a small desk, (d) picking up and carrying a 5\,kg loaded basket, (e) asymmetrically carrying a 10\,kg water bottle, (f) walking with two 12\,kg kettlebells, and (g) squatting with the same 24\,kg total payload.}
\label{fig:real_robot_tasks}
\end{figure}

Finally, we deploy the same policy on L7 hardware using only the deployable observation interface. We do not switch policies or task-specific controllers between tracking, object pickup, carrying, squatting, and pushing. As shown in Fig.~\ref{fig:teaser} and Fig.~\ref{fig:real_robot_tasks}, HEFT completes multi-stage object handling, asymmetric carrying, 24\,kg kettlebell walking, 24\,kg kettlebell squatting, backpack pickup and placement, desk lifting, and rack pushing with a single policy. These tasks cover ground-object pickup, coordinated placement, changing payload distribution, maximum-payload walking, large vertical motion under load, bulky two-hand transport, and sustained horizontal pushing. The hardware trials therefore test both sides of the training design: PMG keeps the controller deployable from raw VR references, while WPC lets the same whole-body mimic policy tolerate large real payloads and contact-rich teleoperation behaviors.

\section{Conclusion}
\label{sec:conclusion}

We presented HEFT, a full-size humanoid teleoperation framework that brings VR tracking and heavy-payload whole-body control into a single policy. HEFT uses reconstructed motion and expert payload labels only during training: PMG lets the controller learn from raw teleoperation references without online reconstruction, while WPC replaces a global payload schedule with motion-dependent payload caps. Experiments show that PMG reduces raw-reference drift on both G1 and L7, and that WPC improves high-payload success while preserving the same deployment interface. On L7 hardware, the resulting policy demonstrates robust payload-bearing teleoperation with real two-hand loads up to 24\,kg.

\section{Limitations}
\label{sec:limitations}

HEFT still requires offline reconstructed VR references and expert rollout labels for training, so transferring to a new robot, tracker setup, or substantially different motion library requires rerunning the data-preparation and WPC labeling pipeline. The current payload model approximates object interaction mainly as wrist-applied forces, and does not explicitly model grasp quality, object geometry, sliding, bracing, or environment-supported contacts. Finally, hardware validation is limited to one full-size platform; evaluating HEFT on more platforms will better test its generality.


\clearpage
\bibliography{example}

\begin{thebibliography}{41}
\providecommand{\natexlab}[1]{#1}
\providecommand{\url}[1]{\texttt{#1}}
\expandafter\ifx\csname urlstyle\endcsname\relax
  \providecommand{\doi}[1]{doi: #1}\else
  \providecommand{\doi}{doi: \begingroup \urlstyle{rm}\Url}\fi

\bibitem[Mandlekar et~al.(2018)Mandlekar, Zhu, Garg, Booher, Spero, Tung, Gao, Emmons, Gupta, Orbay, et~al.]{mandlekar2018roboturk}
A.~Mandlekar, Y.~Zhu, A.~Garg, J.~Booher, M.~Spero, A.~Tung, J.~Gao, J.~Emmons, A.~Gupta, E.~Orbay, et~al.
\newblock Roboturk: A crowdsourcing platform for robotic skill learning through imitation.
\newblock In \emph{Conference on Robot Learning}, pages 879--893. PMLR, 2018.

\bibitem[Zhao et~al.(2023)Zhao, Kumar, Levine, and Finn]{zhao2023learning}
T.~Z. Zhao, V.~Kumar, S.~Levine, and C.~Finn.
\newblock Learning fine-grained bimanual manipulation with low-cost hardware.
\newblock \emph{arXiv preprint arXiv:2304.13705}, 2023.

\bibitem[Fu et~al.(2024)Fu, Zhao, and Finn]{fu2024mobile}
Z.~Fu, T.~Z. Zhao, and C.~Finn.
\newblock Mobile aloha: Learning bimanual mobile manipulation with low-cost whole-body teleoperation.
\newblock \emph{arXiv preprint arXiv:2401.02117}, 2024.

\bibitem[Wu et~al.(2024)Wu, Shentu, Yi, Lin, and Abbeel]{wu2024gello}
P.~Wu, Y.~Shentu, Z.~Yi, X.~Lin, and P.~Abbeel.
\newblock Gello: A general, low-cost, and intuitive teleoperation framework for robot manipulators.
\newblock In \emph{2024 IEEE/RSJ International Conference on Intelligent Robots and Systems (IROS)}, pages 12156--12163. IEEE, 2024.

\bibitem[Cheng et~al.(2024)Cheng, Li, Yang, Yang, and Wang]{cheng2024open}
X.~Cheng, J.~Li, S.~Yang, G.~Yang, and X.~Wang.
\newblock Open-television: Teleoperation with immersive active visual feedback.
\newblock \emph{arXiv preprint arXiv:2407.01512}, 2024.

\bibitem[He et~al.(2024{\natexlab{a}})He, Luo, Xiao, Zhang, Kitani, Liu, and Shi]{he2024learning}
T.~He, Z.~Luo, W.~Xiao, C.~Zhang, K.~Kitani, C.~Liu, and G.~Shi.
\newblock Learning human-to-humanoid real-time whole-body teleoperation.
\newblock In \emph{2024 IEEE/RSJ International Conference on Intelligent Robots and Systems (IROS)}, pages 8944--8951. IEEE, 2024{\natexlab{a}}.

\bibitem[He et~al.(2024{\natexlab{b}})He, Luo, He, Xiao, Zhang, Zhang, Kitani, Liu, and Shi]{he2024omnih2o}
T.~He, Z.~Luo, X.~He, W.~Xiao, C.~Zhang, W.~Zhang, K.~Kitani, C.~Liu, and G.~Shi.
\newblock Omnih2o: Universal and dexterous human-to-humanoid whole-body teleoperation and learning.
\newblock \emph{arXiv preprint arXiv:2406.08858}, 2024{\natexlab{b}}.

\bibitem[Fu et~al.(2024)Fu, Zhao, Wu, Wetzstein, and Finn]{fu2024humanplus}
Z.~Fu, Q.~Zhao, Q.~Wu, G.~Wetzstein, and C.~Finn.
\newblock Humanplus: Humanoid shadowing and imitation from humans.
\newblock \emph{arXiv preprint arXiv:2406.10454}, 2024.

\bibitem[Ze et~al.(2025{\natexlab{a}})Ze, Chen, Ara{\'u}jo, Cao, Peng, Wu, and Liu]{ze2025twist}
Y.~Ze, Z.~Chen, J.~P. Ara{\'u}jo, Z.-a. Cao, X.~B. Peng, J.~Wu, and C.~K. Liu.
\newblock Twist: Teleoperated whole-body imitation system.
\newblock \emph{arXiv preprint arXiv:2505.02833}, 2025{\natexlab{a}}.

\bibitem[Ze et~al.(2025{\natexlab{b}})Ze, Zhao, Wang, Kanazawa, Duan, Abbeel, Shi, Wu, and Liu]{ze2025twist2}
Y.~Ze, S.~Zhao, W.~Wang, A.~Kanazawa, R.~Duan, P.~Abbeel, G.~Shi, J.~Wu, and C.~K. Liu.
\newblock Twist2: Scalable, portable, and holistic humanoid data collection system.
\newblock \emph{arXiv preprint arXiv:2511.02832}, 2025{\natexlab{b}}.

\bibitem[Jiang et~al.(2022)Jiang, Streli, Qiu, Fender, Laich, Snape, and Holz]{jiang2022avatarposer}
J.~Jiang, P.~Streli, H.~Qiu, A.~Fender, L.~Laich, P.~Snape, and C.~Holz.
\newblock Avatarposer: Articulated full-body pose tracking from sparse motion sensing.
\newblock In \emph{European conference on computer vision}, pages 443--460. Springer, 2022.

\bibitem[Ponton et~al.(2023)Ponton, Yun, Aristidou, Andujar, and Pelechano]{ponton2023sparseposer}
J.~L. Ponton, H.~Yun, A.~Aristidou, C.~Andujar, and N.~Pelechano.
\newblock Sparseposer: Real-time full-body motion reconstruction from sparse data.
\newblock \emph{ACM Transactions on Graphics}, 43\penalty0 (1):\penalty0 1--14, 2023.

\bibitem[Winkler et~al.(2022)Winkler, Won, and Ye]{winkler2022questsim}
A.~Winkler, J.~Won, and Y.~Ye.
\newblock Questsim: Human motion tracking from sparse sensors with simulated avatars.
\newblock In \emph{SIGGRAPH Asia 2022 conference papers}, pages 1--8, 2022.

\bibitem[Li et~al.(2023)Li, Liu, and Wu]{li2023ego}
J.~Li, K.~Liu, and J.~Wu.
\newblock Ego-body pose estimation via ego-head pose estimation.
\newblock In \emph{Proceedings of the IEEE/CVF Conference on Computer Vision and Pattern Recognition}, pages 17142--17151, 2023.

\bibitem[Zhang et~al.(2024)Zhang, Bhatnagar, Xu, Winkler, Kadlecek, Tang, and Bogo]{zhang2024rohm}
S.~Zhang, B.~L. Bhatnagar, Y.~Xu, A.~Winkler, P.~Kadlecek, S.~Tang, and F.~Bogo.
\newblock Rohm: Robust human motion reconstruction via diffusion.
\newblock In \emph{Proceedings of the IEEE/CVF Conference on Computer Vision and Pattern Recognition}, pages 14606--14617, 2024.

\bibitem[Rempe et~al.(2021)Rempe, Birdal, Hertzmann, Yang, Sridhar, and Guibas]{rempe2021humor}
D.~Rempe, T.~Birdal, A.~Hertzmann, J.~Yang, S.~Sridhar, and L.~J. Guibas.
\newblock Humor: 3d human motion model for robust pose estimation.
\newblock In \emph{Proceedings of the IEEE/CVF international conference on computer vision}, pages 11488--11499, 2021.

\bibitem[Shin et~al.(2024)Shin, Kim, Halilaj, and Black]{shin2024wham}
S.~Shin, J.~Kim, E.~Halilaj, and M.~J. Black.
\newblock Wham: Reconstructing world-grounded humans with accurate 3d motion.
\newblock In \emph{Proceedings of the IEEE/CVF Conference on Computer Vision and Pattern Recognition}, pages 2070--2080, 2024.

\bibitem[Wang et~al.(2024)Wang, Wang, Liu, and Daniilidis]{wang2024tram}
Y.~Wang, Z.~Wang, L.~Liu, and K.~Daniilidis.
\newblock Tram: Global trajectory and motion of 3d humans from in-the-wild videos.
\newblock In \emph{European Conference on Computer Vision}, pages 467--487. Springer, 2024.

\bibitem[Shen et~al.(2024)Shen, Pi, Xia, Cen, Peng, Hu, Bao, Hu, and Zhou]{shen2024world}
Z.~Shen, H.~Pi, Y.~Xia, Z.~Cen, S.~Peng, Z.~Hu, H.~Bao, R.~Hu, and X.~Zhou.
\newblock World-grounded human motion recovery via gravity-view coordinates.
\newblock In \emph{SIGGRAPH Asia 2024 Conference Papers}, pages 1--11, 2024.

\bibitem[Dao et~al.(2024)Dao, Duan, and Fern]{dao2024sim}
J.~Dao, H.~Duan, and A.~Fern.
\newblock Sim-to-real learning for humanoid box loco-manipulation.
\newblock In \emph{2024 IEEE International Conference on Robotics and Automation (ICRA)}, pages 16930--16936. IEEE, 2024.

\bibitem[Zhang et~al.(2025)Zhang, Yuan, Gurunath, Gupta, Omidshafiei, Agha-mohammadi, Vazquez-Chanlatte, Pedersen, He, and Shi]{zhang2025falcon}
Y.~Zhang, Y.~Yuan, P.~Gurunath, I.~Gupta, S.~Omidshafiei, A.-a. Agha-mohammadi, M.~Vazquez-Chanlatte, L.~Pedersen, T.~He, and G.~Shi.
\newblock Falcon: Learning force-adaptive humanoid loco-manipulation.
\newblock \emph{arXiv preprint arXiv:2505.06776}, 2025.

\bibitem[Purushottam et~al.(2025)Purushottam, Yan, Xu, and Ramos]{purushottam2025heavy}
A.~Purushottam, J.~Yan, C.~Xu, and J.~Ramos.
\newblock Heavy lifting tasks via haptic teleoperation of a wheeled humanoid.
\newblock In \emph{2025 IEEE-RAS 24th International Conference on Humanoid Robots (Humanoids)}, pages 345--350. IEEE, 2025.

\bibitem[Wang et~al.(2026)Wang, Zhang, Xie, Yu, Song, Bai, and Zhu]{wang2026halo}
X.~Wang, C.~Zhang, W.~Xie, C.~Yu, W.~Song, C.~Bai, and S.~Zhu.
\newblock Halo: Closing sim-to-real gap for heavy-loaded humanoid agile motion skills via differentiable simulation.
\newblock \emph{arXiv preprint arXiv:2603.15084}, 2026.

\bibitem[Peng et~al.(2018)Peng, Abbeel, Levine, and Van~de Panne]{peng2018deepmimic}
X.~B. Peng, P.~Abbeel, S.~Levine, and M.~Van~de Panne.
\newblock Deepmimic: Example-guided deep reinforcement learning of physics-based character skills.
\newblock \emph{ACM Transactions On Graphics (TOG)}, 37\penalty0 (4):\penalty0 1--14, 2018.

\bibitem[Peng et~al.(2021)Peng, Ma, Abbeel, Levine, and Kanazawa]{peng2021amp}
X.~B. Peng, Z.~Ma, P.~Abbeel, S.~Levine, and A.~Kanazawa.
\newblock Amp: Adversarial motion priors for stylized physics-based character control.
\newblock \emph{ACM Transactions on Graphics (ToG)}, 40\penalty0 (4):\penalty0 1--20, 2021.

\bibitem[Peng et~al.(2022)Peng, Guo, Halper, Levine, and Fidler]{peng2022ase}
X.~B. Peng, Y.~Guo, L.~Halper, S.~Levine, and S.~Fidler.
\newblock Ase: Large-scale reusable adversarial skill embeddings for physically simulated characters.
\newblock \emph{ACM Transactions On Graphics (TOG)}, 41\penalty0 (4):\penalty0 1--17, 2022.

\bibitem[Luo et~al.(2023)Luo, Cao, Kitani, Xu, et~al.]{luo2023perpetual}
Z.~Luo, J.~Cao, K.~Kitani, W.~Xu, et~al.
\newblock Perpetual humanoid control for real-time simulated avatars.
\newblock In \emph{Proceedings of the IEEE/CVF International Conference on Computer Vision}, pages 10895--10904, 2023.

\bibitem[Luo et~al.(2024)Luo, Cao, Merel, Winkler, Huang, Kitani, and Xu]{luo2024universal}
Z.~Luo, J.~Cao, J.~Merel, A.~Winkler, J.~Huang, K.~Kitani, and W.~Xu.
\newblock Universal humanoid motion representations for physics-based control.
\newblock In \emph{International Conference on Learning Representations}, volume 2024, pages 56766--56782, 2024.

\bibitem[Tessler et~al.(2024)Tessler, Guo, Nabati, Chechik, and Peng]{tessler2024maskedmimic}
C.~Tessler, Y.~Guo, O.~Nabati, G.~Chechik, and X.~B. Peng.
\newblock Maskedmimic: Unified physics-based character control through masked motion inpainting.
\newblock \emph{ACM Transactions On Graphics (TOG)}, 43\penalty0 (6):\penalty0 1--21, 2024.

\bibitem[Liao et~al.(2025)Liao, Truong, Huang, Gao, Tevet, Sreenath, and Liu]{liao2025beyondmimic}
Q.~Liao, T.~E. Truong, X.~Huang, Y.~Gao, G.~Tevet, K.~Sreenath, and C.~K. Liu.
\newblock Beyondmimic: From motion tracking to versatile humanoid control via guided diffusion.
\newblock \emph{arXiv preprint arXiv:2508.08241}, 2025.

\bibitem[Yin et~al.(2026)Yin, Zeng, Fan, Dai, Wang, Zhang, Tian, Wang, Pang, and Zhang]{yin2026unitracker}
K.~Yin, W.~Zeng, K.~Fan, M.~Dai, Z.~Wang, Q.~Zhang, Z.~Tian, J.~Wang, J.~Pang, and W.~Zhang.
\newblock Unitracker: Learning universal whole-body motion tracker for humanoid robots.
\newblock \emph{IEEE Robotics and Automation Letters}, 2026.

\bibitem[Luo et~al.(2026)Luo, Yuan, Wang, Li, Castañeda, Chen, Cao, Li, Minor, Ben, Park, Sami, Wang, Da, Ding, Hogg, Song, Lim, Jeong, He, Xue, Xiao, Yuen, Kautz, Chang, Iqbal, Fan, and Zhu]{luo2026sonic}
Z.~Luo, Y.~Yuan, T.~Wang, C.~Li, F.~Castañeda, S.~Chen, Z.-A. Cao, J.~Li, D.~Minor, Q.~Ben, J.~Park, D.~Sami, Z.~Wang, X.~Da, R.~Ding, C.~Hogg, L.~Song, E.~Lim, E.~Jeong, T.~He, H.~Xue, W.~Xiao, S.~Yuen, J.~Kautz, Y.~Chang, U.~Iqbal, L.~J. Fan, and Y.~Zhu.
\newblock Sonic: Supersizing motion tracking for natural humanoid whole-body control, 2026.
\newblock URL \url{https://arxiv.org/abs/2511.07820}.

\bibitem[Mahmood et~al.(2019)Mahmood, Ghorbani, Troje, Pons-Moll, and Black]{mahmood2019amass}
N.~Mahmood, N.~Ghorbani, N.~F. Troje, G.~Pons-Moll, and M.~J. Black.
\newblock Amass: Archive of motion capture as surface shapes.
\newblock In \emph{Proceedings of the IEEE/CVF international conference on computer vision}, pages 5442--5451, 2019.

\bibitem[Fu et~al.(2023)Fu, Cheng, and Pathak]{fu2023deep}
Z.~Fu, X.~Cheng, and D.~Pathak.
\newblock Deep whole-body control: learning a unified policy for manipulation and locomotion.
\newblock In \emph{Conference on Robot Learning}, pages 138--149. PMLR, 2023.

\bibitem[Rigo et~al.(2024)Rigo, Hu, Gupta, and Nguyen]{rigo2024hierarchical}
A.~Rigo, M.~Hu, S.~K. Gupta, and Q.~Nguyen.
\newblock Hierarchical optimization-based control for whole-body loco-manipulation of heavy objects.
\newblock In \emph{2024 IEEE International Conference on Robotics and Automation (ICRA)}, pages 15322--15328. IEEE, 2024.

\bibitem[{Bones Studio}(2026)]{bonesseed}
{Bones Studio}.
\newblock {BONES-SEED}: Skeletal everyday embodiment dataset, 2026.
\newblock Motion data by Bones Studio, available at https://bones.studio/datasets/seed.

\bibitem[Mason et~al.(2022)Mason, Starke, and Komura]{mason2022real}
I.~Mason, S.~Starke, and T.~Komura.
\newblock Real-time style modelling of human locomotion via feature-wise transformations and local motion phases.
\newblock \emph{Proceedings of the ACM on Computer Graphics and Interactive Techniques}, 5\penalty0 (1):\penalty0 1--18, 2022.

\bibitem[Harvey et~al.(2020)Harvey, Yurick, Nowrouzezahrai, and Pal]{harvey2020robust}
F.~G. Harvey, M.~Yurick, D.~Nowrouzezahrai, and C.~Pal.
\newblock Robust motion in-betweening.
\newblock 39\penalty0 (4), 2020.

\bibitem[Pavlakos et~al.(2019)Pavlakos, Choutas, Ghorbani, Bolkart, Osman, Tzionas, and Black]{pavlakos2019expressive}
G.~Pavlakos, V.~Choutas, N.~Ghorbani, T.~Bolkart, A.~A. Osman, D.~Tzionas, and M.~J. Black.
\newblock Expressive body capture: 3d hands, face, and body from a single image.
\newblock In \emph{Proceedings of the IEEE/CVF conference on computer vision and pattern recognition}, pages 10975--10985, 2019.

\bibitem[Kumar et~al.(2021)Kumar, Fu, Pathak, and Malik]{kumar2021rma}
A.~Kumar, Z.~Fu, D.~Pathak, and J.~Malik.
\newblock Rma: Rapid motor adaptation for legged robots.
\newblock \emph{arXiv preprint arXiv:2107.04034}, 2021.

\bibitem[Schulman et~al.(2017)Schulman, Wolski, Dhariwal, Radford, and Klimov]{schulman2017proximal}
J.~Schulman, F.~Wolski, P.~Dhariwal, A.~Radford, and O.~Klimov.
\newblock Proximal policy optimization algorithms.
\newblock \emph{arXiv preprint arXiv:1707.06347}, 2017.

\end{thebibliography}

\clearpage
\appendix
\compactappendixspacing
\numberwithin{table}{section}
\numberwithin{figure}{section}
\numberwithin{equation}{section}

\section{Reference Datasets}
\label{app:reference_data}

This section documents the reference streams used throughout training and evaluation. Table~\ref{tab:app_reference_data} summarizes the mocap libraries, paired VR set, and held-out evaluation splits; Table~\ref{tab:app_dynamic_weights} gives the statistics used to form the dynamic split.

\begin{table}[H]
\centering
\scriptsize
\caption{Reference data and evaluation splits.}
\label{tab:app_reference_data}
\setlength{\tabcolsep}{2.5pt}
\renewcommand{\arraystretch}{1.05}
\begin{tabular*}{\linewidth}{@{\extracolsep{\fill}}lrrl@{}}
\toprule
\textbf{Set} & \textbf{Clips} & \textbf{Duration} & \textbf{Source} \\
\midrule
SEED train library & 61,612 & 134.35\,h & SEED \\
100STYLE train library & 810 & 18.78\,h & 100STYLE \\
LaFAN1 train library & 40 & 2.19h & LaFAN1 \\
Paired VR training set & 643 & 6.08\,h & VR teleoperation \\
\midrule
$\mathcal{D}_{\mathrm{VR}}$ & 8 & 52.26\,s & Held-out VR \\
$\mathcal{D}_{\mathrm{random}}$ & 100 & 21.62\,min & Held-out SEED \\
$\mathcal{D}_{\mathrm{dynamic}}$ & 100 & 8.23\,min & Held-out SEED \\
\bottomrule
\end{tabular*}
\end{table}

\paragraph{$\mathcal{D}_{\mathrm{dynamic}}$ selection.}
The high-dynamic split is selected from the SEED clean clips, using only reference-motion statistics before any policy rollout. All statistics are computed from the 50\,Hz retargeted reference. For each candidate clip $i$, we compute a dynamics feature vector $\phi_i$:
\begin{equation}
\begin{split}
\phi_i=\{&
\bar v_{xy},\, v^{\max}_{xy},\, v^{95}_{\mathrm{root}},\, a^{95}_{\mathrm{root}},\,
\Delta z,\, v^{\max}_{z},\,
\mathrm{rms}(\dot\psi),\, \dot\psi^{95},\\
&\mathrm{rms}(\dot q),\, \dot q^{95},\,
\mathrm{rms}(\ddot q),\, \ddot q^{95}\}.
\end{split}
\end{equation}
Here $v_{xy}$ is horizontal root speed, $v_{\mathrm{root}}$ and $a_{\mathrm{root}}$ are 3D root speed and acceleration, $\Delta z$ is root-height range, $\dot\psi$ is root yaw rate, and $q$ denotes retargeted robot joints. RMS and 95th-percentile terms are computed over frames. Each feature is percentile-normalized over the candidate pool by
\begin{equation}
  \tilde\phi_{i,k}=
  \mathrm{clip}\left(
  \frac{\phi_{i,k}-P_5(\phi_{\cdot,k})}{P_{95}(\phi_{\cdot,k})-P_5(\phi_{\cdot,k})},
  0,1
  \right),
\end{equation}
where $P_5$ and $P_{95}$ are percentiles over all candidate clips. The dynamic score is $d_i=\sum_k w_k\tilde\phi_{i,k}$, using the weights in Table~\ref{tab:app_dynamic_weights}.

\begin{table}[H]
\centering
\scriptsize
\caption{Feature weights for selecting $\mathcal{D}_{\mathrm{dynamic}}$.}
\label{tab:app_dynamic_weights}
\setlength{\tabcolsep}{3pt}
\setlength{\extrarowheight}{1pt}
\renewcommand{\arraystretch}{1.18}
\begin{tabular*}{\linewidth}{@{\extracolsep{\fill}}lll@{}}
\toprule
\textbf{Feature group} & \textbf{Statistic} & \textbf{Weight} \\
\midrule
Horizontal root speed & $\bar v_{xy}$, $v^{\max}_{xy}$ & 0.08, 0.08 \\
Root speed/acceleration & $v^{95}_{\mathrm{root}}$, $a^{95}_{\mathrm{root}}$ & 0.10, 0.12 \\
Vertical motion & $\Delta z$, $v^{\max}_{z}$ & 0.10, 0.10 \\
Yaw motion & $\mathrm{rms}(\dot\psi)$, $\dot\psi^{95}$ & 0.08, 0.08 \\
Joint velocity & $\mathrm{rms}(\dot q)$, $\dot q^{95}$ & 0.10, 0.08 \\
Joint acceleration & $\mathrm{rms}(\ddot q)$, $\ddot q^{95}$ & 0.10, 0.08 \\
\bottomrule
\end{tabular*}
\end{table}
We sort all candidates by $d_i$ in descending order and take the top 100 clips. Repeated actions are retained if their reference statistics rank highly.

\section{Privileged Motion Guidance Details}
\label{app:pmg_details}

This section specifies the PMG observation interface. The policy observes a short motion window rather than only the current target pose, which provides local motion context for tracking. We describe this window with reference steps, defined as integer frame offsets from the current 50\,Hz control step. Positive offsets provide future reference targets, while negative offsets provide reference history. Table~\ref{tab:app_actor_obs} lists the deployable actor observation; for L7 and G1, $n_q=29$ denotes the joint dimension.

With this notation, we use different reference windows for the teacher and the student. During training, the teacher can use a longer look-ahead because it does not affect deployment latency. At deployment, future student targets require buffering the online VR stream, so longer look-ahead directly increases teleoperation delay. We therefore use a shorter student reference window with both future and history steps, together with a proprioceptive history window. We denote the numbers of teacher reference steps, student reference steps, and proprioceptive history steps by $H_t$, $H_s$, and $H_p$, respectively.

\begin{table}[H]
\centering
\scriptsize
\caption{Actor observation for the deployable student policy.}
\label{tab:app_actor_obs}
\setlength{\tabcolsep}{2.2pt}
\renewcommand{\arraystretch}{1.02}
\begin{tabularx}{\linewidth}{@{}lYcc@{}}
\toprule
\textbf{Signal} & \textbf{Definition} & \textbf{Shape}  \\
\midrule
Root position command & future root-position offsets from $S_{\mathrm{raw}}$, in the reference root frame; current frame excluded & $(H_{t,s}-1)\times3$  \\
Root orientation command & root orientation difference from $S_{\mathrm{raw}}$, represented as 6D rotation & $H_{t,s}\times6$  \\
Target joints & retargeted joint positions from $S_{\mathrm{raw}}$ & $H_{t,s}\times n_q$  \\
Target joint difference & raw-reference joint target minus measured joint position & $H_{t,s}\times n_q$ \\
Target root height & root $z$ from $S_{\mathrm{raw}}$ & $H_{t,s}\times1$  \\
Target gravity & projected gravity from raw-reference root orientation & $H_{t,s}\times3$  \\
Root angular velocity & measured IMU angular velocity history & $H_p\times3$  \\
Projected gravity & measured IMU gravity history & $H_p\times3$ \\
Joint position history & encoder joint-position history & $H_p\times n_q$  \\
Joint velocity history & encoder joint-velocity history & $H_p\times n_q$  \\
Previous actions & previous joint-position commands & $8\times n_q$  \\
\bottomrule
\end{tabularx}
\end{table}

PMG is implemented through teacher and student motion streams in the motion dataset loader. For paired VR clips, the teacher stream loads the reconstructed and retargeted sequence, while the student stream loads the corresponding raw VR-derived sequence. For ordinary mocap clips, no separate student stream is provided, so the student reference falls back to the teacher reference. The loader checks that paired teacher/student clips have matched joint names, body names, motion counts, and sequence lengths before constructing the mapping. At each environment step, the command module retrieves both motion slices from the dataloader: a student-horizon slice from the student stream for constructing policy observations, and a teacher-horizon slice from the teacher stream for privileged observations and reward computation.

\section{Windowed Payload Curriculum Details}
\label{app:wpc_details}

\begin{figure}[H]
\centering
\includegraphics[width=0.9\linewidth]{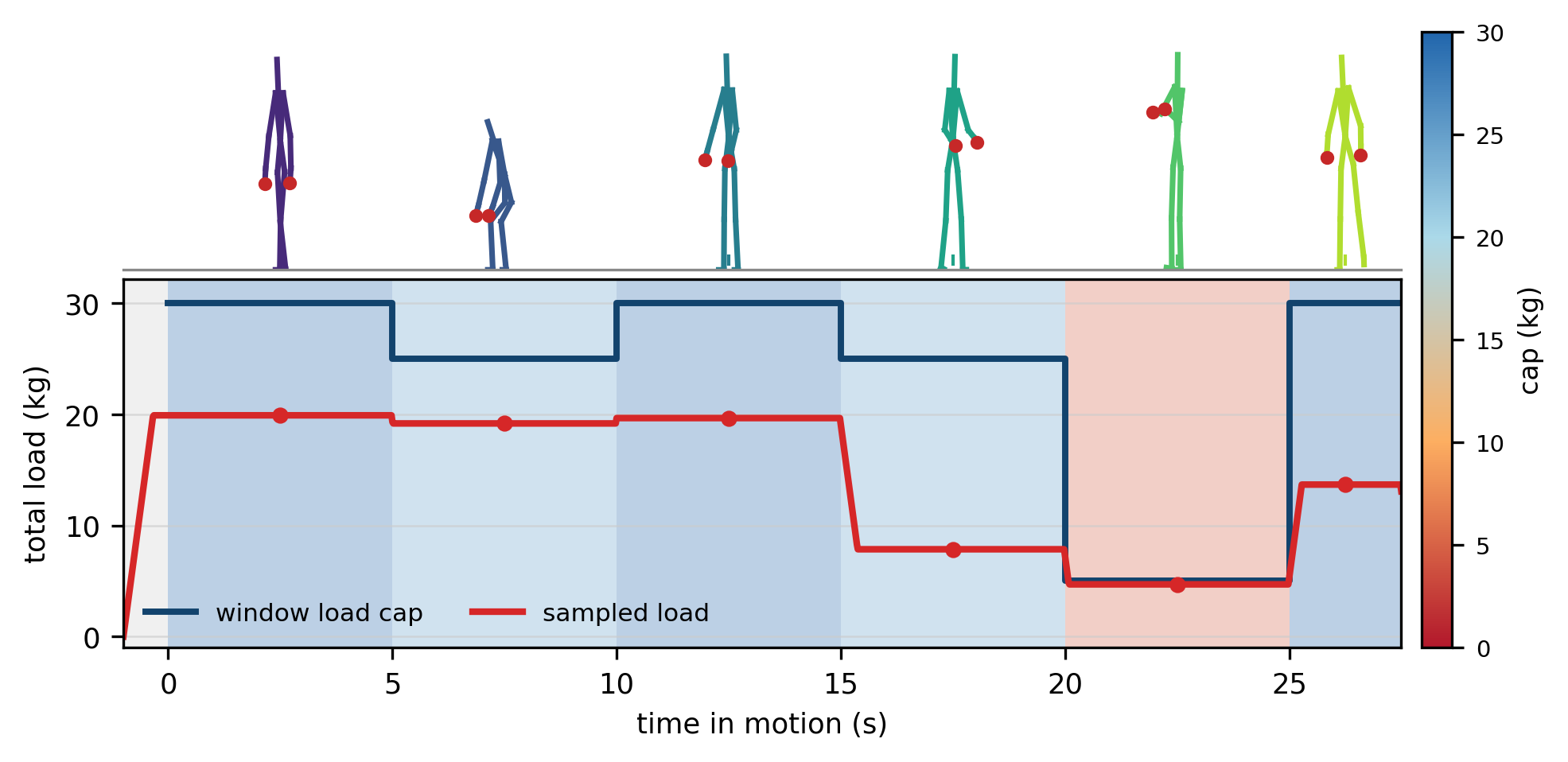}
\caption{Example WPC label and sampled load trace. The feasible payload cap changes with motion window, so the curriculum samples high load only where the expert-labeled window is feasible.}
\label{fig:app_wpc_cap_trace}
\end{figure}

This section explains how WPC converts clean-reference expert rollouts into motion-window payload limits. Algorithm~\ref{alg:wpc} gives the 5\,s window search, and Fig.~\ref{fig:app_wpc_cap_trace} shows how the labels cap sampled payloads during final training.

\paragraph{Expert labeling.}
The WPC expert is used only to label payload feasibility. It tracks the clean stream for both mocap and paired VR clips, with observation noise and domain randomization disabled during labeling. Algorithm~\ref{alg:wpc} describes how the expert assigns a cap to each 5\,s motion window; the blue curve in Fig.~\ref{fig:app_wpc_cap_trace} shows an example cap trace.

\begin{algorithm}[H]
\caption{Windowed payload-cap labeling}
\label{alg:wpc}
\footnotesize
\begin{algorithmic}[1]
\Require expert $\pi_E$; motion set $\mathcal{M}$; window length $T_w=5$\,s; load grid $\{30,25,\ldots,0\}$\,kg
\Ensure window caps $\{\bar m_w\}$ for WPC sampling
\For{each motion $m_i\in\mathcal{M}$}
    \State Split $m_i$ into non-overlapping windows $w_{i,k}$ of length $T_w$
    \For{each window $w_{i,k}$}
        \State $\bar m_{w_{i,k}}\gets 0$
        \For{each candidate load $c\in\{30,25,\ldots,0\}$\,kg}
            \State Roll out $\pi_E$ from the 1\,s ramp start through the end of $w_{i,k}$ under load $c$
            \If{the rollout completes with no termination}
                \State $\bar m_{w_{i,k}}\gets c$
                \State \textbf{break}
            \EndIf
        \EndFor
    \EndFor
\EndFor
\end{algorithmic}
\end{algorithm}

\paragraph{Training sampler and force model.}
During final policy training, the active window index $w(t)$ selects the cap $\bar m_{w(t)}$, and the curriculum samples a target total payload inside that cap. With training progress $p$, the load scale ramps linearly to full strength by 80\% training progress.
 The red curve in Fig.~\ref{fig:app_wpc_cap_trace} shows the sampled payload remaining below the active cap. The total wrist force is applied to the left and right wrist-roll links with random split. The force direction is sampled inside a 12$^\circ$ cone around the commanded payload direction.

\section{Training and Deployment Details}
\label{app:training_deployment}

This section records implementation details. Table~\ref{tab:app_network} gives the network modules, Table~\ref{tab:app_training_hparams} gives the PPO budget, Table~\ref{tab:app_reward_terms} defines rewards.

\paragraph{Policy networks.}
The teacher and student actors share the same MLP backbone size but condition on different latent inputs: the teacher actor receives the privileged latent $z_t$ from the privileged encoder, while the student actor receives the adapter prediction $\hat z_t$. Both latents are 256-D. The actor outputs the mean of a residual joint-position command, with a learned joint-wise diagonal Gaussian standard deviation during training. At deployment, only the adapter and student actor are executed.
\begin{table}[H]
\centering
\scriptsize
\caption{Network architecture.}
\label{tab:app_network}
\setlength{\tabcolsep}{3pt}
\renewcommand{\arraystretch}{1.05}
\begin{tabularx}{\linewidth}{@{}lYl@{}}
\toprule
\textbf{Module} & \textbf{Input} & \textbf{Architecture} \\
\midrule
Teacher actor & deployable observation plus privileged latent $z_t$ & MLP $[1024,1024,512]$ to Gaussian joint-action head. \\
Student actor & deployable observation plus adapter latent $\hat z_t$ & MLP $[1024,1024,512]$ to Gaussian joint-action head. \\
Critic & deployable observation, privileged observation, critic-privileged variables & MLP $[1024,512,512]$ to scalar value. \\
Privileged encoder & clean-reference and simulator privileged observation & MLP $[512]$ to 256-D privileged latent. \\
Adapter / RMA encoder & deployable observation only & MLP $[1024,512]$ to 256-D latent prediction. \\
Action distribution & actor mean and learned standard deviation & Diagonal Gaussian; joint-specific initialization noise scale. \\
\bottomrule
\end{tabularx}
\end{table}

\begin{table}[H]
\centering
\scriptsize
\caption{Training hyperparameters.}
\label{tab:app_training_hparams}
\setlength{\tabcolsep}{3pt}
\renewcommand{\arraystretch}{1.05}
\begin{tabular*}{0.4\linewidth}{@{\extracolsep{\fill}}ll@{}}
\toprule
\textbf{Parameter} & \textbf{Value} \\
\midrule
Parallel environments & $8{\times} 8192$ \\
Physics / control step & 0.005\,s / 0.020\,s \\
Rollout length & 32 steps \\
PPO epochs / minibatches & 3 / 8 \\
Discount / GAE & 0.99 / 0.95 \\
PPO clip & 0.2 \\
Adapter loss weight & 0.2 \\
Training budget & $5{\times}10^9$ frames \\
\bottomrule
\end{tabular*}
\end{table}
\paragraph{Reward computation.}
For all tracking terms, the reward is computed as the form:
\begin{equation}
  r_k = w_k \exp\left(-e_k / \sigma_k\right),
\end{equation}
where $e_k$ is the scalar tracking error in Table~\ref{tab:app_reward_terms}.
The keypoint rewards use a sparse set of 13 representative rigid bodies on both L7 and G1, covering the head, hands, selected upper-limb links, hip-yaw links, knee links, and ankle-roll links.

\begin{table}[H]
\centering
\scriptsize
\caption{Reward terms for training. Tracking targets use $S_{\mathrm{clean}}$ for paired VR clips.}
\label{tab:app_reward_terms}
\setlength{\tabcolsep}{2.2pt}
\renewcommand{\arraystretch}{1.02}
\begin{tabularx}{\linewidth}{@{}lccY@{}}
\toprule
\textbf{Term} & \textbf{$w$} & \textbf{$\sigma$} & \textbf{Unweighted term / error definition} \\
\midrule
Root position & 0.5 & 0.3 & $e=\lVert p_{\mathrm{root}}-\hat p_{\mathrm{root}}\rVert_2$. \\
Root rotation & 0.5 & 0.4 & $e=\lVert\mathrm{Log}(q_{\mathrm{root}}^{-1}\hat q_{\mathrm{root}})\rVert_2$. \\
Root linear velocity & 1.0 & 1.0 & $e=\lVert R^\top v_{\mathrm{root}}-\hat R^\top \hat v_{\mathrm{root}}\rVert_2$. \\
Root angular velocity & 1.0 & 3.0 & $e=\lVert R^\top \omega_{\mathrm{root}}-\hat R^\top \hat \omega_{\mathrm{root}}\rVert_2$. \\
Keypoint position & 2.0 & 0.3 & $e=K^{-1}\sum_j\lVert x_j-\hat x_j\rVert_2$, body-frame keypoints. \\
Keypoint linear velocity & 1.0 & 1.0 & $e=K^{-1}\sum_j\lVert \dot x_j-\hat{\dot x}_j\rVert_2$. \\
Keypoint rotation & 2.0 & 0.4 & $e=K^{-1}\sum_j\lVert\mathrm{Log}(q_j^{-1}\hat q_j)\rVert_2$. \\
Keypoint angular velocity & 1.0 & 3.0 & $e=K^{-1}\sum_j\lVert \omega_j-\hat\omega_j\rVert_2$. \\
Joint position & 1.0 & 0.5 & $e=n_q^{-1}\sum_l|q_l-\hat q_l|$. \\
Joint velocity & 0.5 & 3.0 & $e=n_q^{-1}\sum_l|\dot q_l-\hat{\dot q}_l|$. \\
Survival & 3.0 & -- & $r=1$ while the rollout is not terminated. \\
Joint velocity regularizer & $5{\times}10^{-4}$ & -- & $r=-\sum_l \dot q_l^2$. \\
Action-rate regularizer & 0.02 & -- & $r=-\lVert a_t-a_{t-1}\rVert_2^2$. \\
Reference foot air time & 5.0 & -- & Sparse contact-timing reward using clean-reference foot-contact state. \\
Dense foot-contact reward & 1.0 & -- & Contact mismatch gives $-1$; matched air/contact states receive height-shaped penalties. \\
Joint position limit & 1.0 & -- & Negative normalized violation outside 90\% soft joint-position limits. \\
Joint torque limit & 0.01 & -- & Negative violation outside 75\% actuator torque limits. \\
\bottomrule
\end{tabularx}
\end{table}

\section{Evaluation Protocol}
\label{app:evaluation_results}

This section records the evaluation definitions and comparison setup used by the main experiments. The equations below define the reported simulation metrics, Table~\ref{tab:app_real_robot_audit} gives the hardware audit trail for the real-object demonstrations.

\paragraph{Metrics.}
For $N$ rollouts, success is the fraction of rollouts that complete the requested horizon without early termination:
\begin{equation}
  \mathrm{Success}=\frac{1}{N}\sum_{i=1}^{N}\mathbf{1}[\mathrm{complete}(i)].
\end{equation}
The final horizontal root error compares start-to-end displacement:
\begin{equation}
  e_{\mathrm{XY}}^i =
  \left\|
  \left(p_{i,T}^{\mathrm{root}}-p_{i,0}^{\mathrm{root}}\right)_{xy}
  -
  \left(\hat p_{i,T}^{\mathrm{root}}-\hat p_{i,0}^{\mathrm{root}}\right)_{xy}
  \right\|_2 .
\end{equation}
Mean per-joint position error is computed over Cartesian body keypoints:
\begin{equation}
  \mathrm{MPJPE} =
  \frac{1}{\sum_i T_i J}
  \sum_i\sum_{t=1}^{T_i}\sum_{j=1}^{J}
  \left\|x_{i,t,j}-\hat{x}_{i,t,j}\right\|_2 .
\end{equation}
Root velocity and angular-velocity errors are frame-weighted means in the robot body frame:
\begin{equation}
  e_v =
  \frac{1}{\sum_i T_i}\sum_i\sum_t
  \left\|R_{i,t}^{\top}v_{i,t}^{\mathrm{root}}-\hat R_{i,t}^{\top}\hat v_{i,t}^{\mathrm{root}}\right\|_2,
  \quad
  e_\omega =
  \frac{1}{\sum_i T_i}\sum_i\sum_t
  \left\|R_{i,t}^{\top}\omega_{i,t}^{\mathrm{root}}-\hat R_{i,t}^{\top}\hat\omega_{i,t}^{\mathrm{root}}\right\|_2 .
\end{equation}
Unless stated otherwise, reported means are frame-weighted across all completed and evaluated rollout frames, and success is segment-weighted.

\begin{table}[H]
\centering
\scriptsize
\caption{Real-robot task audit corresponding to Fig.~\ref{fig:real_robot_tasks}.}
\label{tab:app_real_robot_audit}
\setlength{\tabcolsep}{3pt}
\renewcommand{\arraystretch}{1.05}
\begin{tabularx}{\linewidth}{@{}llllY@{}}
\toprule
\textbf{Panel} & \textbf{Task} & \textbf{Object} & \textbf{Mass} & \textbf{Notes} \\
\midrule
(a) & Pickup and place & backpack & 5\,kg & Ground pickup followed by chair placement. \\
(b) & Push & wheeled rack & not measured & Sustained horizontal pushing under teleoperation. \\
(c) & Lift and carry & small desk & 8\,kg & Bulky two-hand object with changing arm posture. \\
(d) & Pickup and carry & loaded basket & 5\,kg & Multi-stage tabletop handling and walking. \\
(e) & Asymmetric carry & water bottle & 10\,kg & Uneven two-hand load distribution. \\
(f) & Walk & two kettlebells & $2\times12$\,kg & Two-hand load during locomotion. \\
(g) & Squat & two kettlebells & $2\times12$\,kg & Large vertical root motion under the same load. \\
\bottomrule
\end{tabularx}
\end{table}

\end{document}